\documentclass{interact}

\usepackage[utf8]{inputenc}
\usepackage[T1]{fontenc}
\usepackage{amsmath,amssymb}
\usepackage{tipa}
\usepackage{graphicx}
\usepackage{booktabs}
\usepackage[natbibapa,nodoi]{apacite}
\usepackage{hyperref}

\begin{document}

\articletype{RESEARCH ARTICLE}

\title{Cross-Linguistic Distinctive-Feature Constraints on Segmental Triphones: Evidence for Contrastive Dispersion}

\author{
\name{Leonardo Ara\'{u}jo\textsuperscript{a}\thanks{CONTACT Leonardo Ara\'{u}jo. Email: leolca@ufsj.edu.br}}
\affil{\textsuperscript{a}Departamento de Engenharia de Telecomunica\c{c}\~{o}es e Mecatr\^{o}nica, Universidade Federal de S\~{a}o Jo\~{a}o del-Rei (UFSJ), Ouro Branco, MG, Brazil}
}

\maketitle

\begin{abstract}
Natural speech reflects articulatory and perceptual constraints, yet how these pressures shape consecutive segment transitions across languages remains underexplored. We analyze adjacent distinctive-feature distances in token-weighted segmental triphones across 17 Universal Dependencies corpora spanning nine language families, transcribed into IPA and mapped to 24 PanPhon features using Hamming and cosine metrics. Natural sequences exhibit an intermediate-distance mode (8--10 features under Hamming, except Vietnamese; 0.28--0.38 under cosine). Against an unrestricted within-word shuffled baseline, natural words reject the inventory-only null ($W=0, p < 10^{-4}$), showing higher expected distance, narrower dispersion, and lower joint triphone entropy in all 17 languages. Under a stricter within-word consonant--vowel template baseline, expected distance and dispersion are no longer separated, whereas joint entropy remains lower in 16 languages ($p = 1.07 \times 10^{-4}$). Entropy decomposition reveals higher mutual information between incoming and outgoing transitions in all 17 languages and all nine families ($W=0, p < 10^{-4}$), demonstrating sequential coupling beyond syllable-template alternation. Word-level mean--SD correlations and directional asymmetries do not replicate cross-linguistically.
\end{abstract}


\section{Introduction}
Phonological systems are shaped by competing functional pressures associated with articulatory economy and perceptual distinctiveness \citep{zipf1949,lindblom1990,flemming2004}. While this trade-off is well established in paradigmatic inventory typology (such as vowel spacing), its quantitative signature in syntagmatic sequencing---across consecutive segments within words---has received comparatively less cross-linguistic scrutiny. It is important to distinguish \emph{paradigmatic dispersion} (how far apart contrasting phonemes sit in an inventory), \emph{sequential feature-space distance} (how much consecutive segments differ in distinctive features), and \emph{syllabic organization} (such as consonant--vowel alternation).

In earlier work on English \citep{araujo2013thesis}, triphones extracted from literary text transcribed via a pronouncing dictionary were analyzed using a discrete distinctive-feature distance, revealing that the joint distribution of adjacent feature distances peaks at intermediate values rather than at extreme similarity or extreme contrast. Focusing on triphones $(\phi_1, \phi_2, \phi_3)$ is motivated on both speech-modeling and analytical grounds. In speech recognition and speech synthesis, triphones model a segment within its immediate left and right phonetic context, capturing short-range co-articulation while scaling more easily across arbitrary vocabularies and typologically diverse languages than full syllable inventories (which grow very large in languages with complex syllable structures). Analytically and statistically, a triphone is the minimal sequential unit that jointly encodes a segment's incoming transition ($d_1$) and outgoing transition ($d_2$) around a central segment $\phi_2$, allowing the joint distribution of consecutive transitions and directional asymmetry to be studied while avoiding the severe data sparsity of longer $n$-grams in moderate-sized corpora.

Here, we scale this distinctive-feature framework to 17 languages across nine language families (nine non-Indo-European and eight Indo-European) using standardized corpora and universal articulatory feature representations, and introduce two nested permutation baselines: an unrestricted within-word shuffle and a consonant--vowel (C/V) template-preserving shuffle. This design isolates whether sequential feature-distance constraints extend beyond both marginal segment inventories and basic C/V alternation. Specifically, we test two nested null hypotheses: ($H_{0,\text{unrest}}$) conditional on each word's segment multiset and token frequency, sequential ordering does not affect $\mathbb{E}[d]$, $\sigma(d)$, or $H(d_1, d_2)$ relative to unrestricted within-word permutations; and ($H_{0,\text{CV}}$) conditional on each word's C/V template, sub-syllabic sequencing does not further constrain $\mathbb{E}[d]$, $\sigma(d)$, or $H(d_1, d_2)$ relative to C/V-template-preserving permutations. Under these controls, we evaluate whether natural sequences exhibit (H1) an intermediate-distance mode, (H2) higher expected adjacent feature distance than unrestricted shuffles, (H3) narrower dispersion and lower joint entropy even relative to C/V-preserving shuffles, and whether earlier single-language findings---(H4) negative within-word mean--SD correlation and (H5) directional asymmetry ($d_1 > d_2$)---survive cross-linguistic and baseline controls.

\section{Literature Review}
\subsection{Articulatory Economy and Perceptual Dispersion}
A central tenet of functional and quantitative linguistics is that phonological systems are governed by a continuous tug-of-war between two opposing communicative forces: \emph{articulatory economy} (for the speaker) and \emph{perceptual distinctiveness} (for the listener). \citet{zipf1949} framed this opposition under the Principle of Least Effort, arguing that speakers naturally favor shorter, less effortful utterances and reduced articulatory gestures, whereas listeners require sufficiently distinct signals to recover the intended message without ambiguity. \citet{martinet1955} developed this insight into a theory of \emph{phonological economy}, proposing that sound systems evolve toward an equilibrium between minimizing the number of articulatory dimensions (syntagmatic and paradigmatic economy) and maintaining functional margins of safety between contrasting phonemes so that communicative distinctions are not lost.

Early quantitative models formalized this equilibrium in static phoneme inventories. \citet{liljencrants1972} simulated vowel systems by placing vowels in a two- or three-dimensional acoustic formant space and maximizing the sum of inverse-square perceptual distances between all pairs of vowels. Despite its simplicity, this maximal-dispersion model successfully predicted the corner-vowel structure of common \mbox{3-}, \mbox{5-}, and \mbox{7-vowel} inventories (such as /i, a, u/). Subsequent work by \citet{lindblom1990} generalized these ideas into the Hyper- and Hypo-articulation (H\&H) theory: rather than aiming for invariant acoustic targets, speakers dynamically modulate articulatory effort along a continuum from \emph{hypo-speech} (economical, undershot gestures when lexical or contextual predictability is high) to \emph{hyper-speech} (precise, effortful articulation when communicative conditions demand high perceptual clarity).

Within constraint-based phonology, \citet{flemming2004} unified these pressures into \emph{Dispersion Theory}, proposing that phonological grammars evaluate candidate contrast sets using three interacting constraints: (1) maximizing the number of contrasts to support a large lexicon, (2) maximizing perceptual distinctiveness (enforcing minimum-distance thresholds between contrasting sounds), and (3) minimizing articulatory effort. Crucially, because maximal perceptual separation often requires extreme or rapid articulatory movements, the interaction of these constraints favors \emph{intermediate} levels of contrast rather than either minimal or maximal separation. While Dispersion Theory and acoustic-space simulations have been extensively tested on static phoneme inventories (especially vowel spacing), far less quantitative work has examined how this same economy--distinctiveness trade-off constrains \emph{sequential} segment-to-segment transitions in running speech across diverse languages.

\subsection{Distinctive Features and Phonetic Distance Metrics}
In linguistics, a \emph{distinctive feature} is the smallest unit of phonological structure that distinguishes one speech sound (phoneme) from another within a language. Distinctive features correspond to articulatory or acoustic properties---such as vocal-fold vibration, airflow, or tongue position---and are typically grouped into major class, laryngeal, manner, and place features. They are commonly represented with $+$ (presence of a property) or $-$ (absence of a property), such as $[+\text{nasal}]$ versus $[-\text{nasal}]$; for example, the voicing feature $[\pm\text{voice}]$ separates $[\text{p}]$ ($[-\text{voice}]$) from $[\text{b}]$ ($[+\text{voice}]$). Sounds that share a feature form natural classes that pattern together in phonological rules, such as voiceless stops taking aspiration in specific prosodic or syllabic positions. Historically, \citet{trubetzkoy1939} laid the groundwork with his theory of distinctive oppositions, \citet{jakobson1952} developed a binary acoustic distinctive-feature framework, and \citet{chomsky1968} formalized a comprehensive articulatory/phonological feature system in \emph{The Sound Pattern of English} (SPE).

Quantitative and computational phonology have built on these representations to define graded segment-to-segment distances. \citet{kondrak2003} introduced ALINE, using multivalued articulatory features with salience weights for phonetic sequence alignment and cognate identification. More recently, \citet{mortensen2016panphon} developed PanPhon, which maps International Phonetic Alphabet (IPA) segments across languages to uniform 24-dimensional ternary articulatory vectors in $\{-1, 0, +1\}^{24}$. Comparing segments via feature Hamming distance counts discrete feature mismatches directly, whereas $\ell_2$-normalized cosine distance equalizes vector norms across segments that differ in their number of unspecified ($0$) entries---a normalization that has proven effective in cross-linguistic phonetic spelling correction \citep{mortensen2016panphon}. Evaluating both metrics together ensures that observed sequential patterns are not artifacts of feature sparsity.

\subsection{Sequential Phonotactics and Quantitative Controls}
At the sequential level, syllabic and phonotactic organization is classically governed by the Sonority Sequencing Principle \citep{clements1990}, which favors rises in sonority from syllable onset to nucleus and falls toward the coda---inherently alternating consonant-like and vowel-like articulations. From an acoustic perspective, \citet{kawasakifukumori1992} showed that universal phonotactic preferences reflect acoustic modulation across consecutive segments, disfavoring sequences with minimal spectral change. Computational models of phonotactic learning further demonstrate that speakers induce gradient, constraint-based generalizations over natural classes of features rather than memorizing raw segment $n$-grams \citep{hayes2008}, while corpus-based diachronic studies show that high functional load helps preserve phonological contrasts against merger \citep{wedel2013}. At the lexical level, \citet{dautriche2017} showed across multiple languages that words exhibit structured phonetic clustering beyond baseline phonotactic regularities, highlighting the importance of comparing natural lexicons against controlled null models. In earlier single-language work on English literary text, \citet{araujo2013thesis} observed that token-weighted triphone intradistances peak at intermediate distinctive-feature distances, that word-level mean and standard deviation of inter-phone distances are negatively correlated, and that triphone feature variability appears asymmetric ($d_1 > d_2$). However, without cross-linguistic replication or a within-word permutation baseline, it remained open which of those patterns reflect universal phonotactic constraints and which arise from inventory-level or combinatorial properties. The present study addresses this gap directly.

\section{Methodology}
\subsection{Corpora and Phonemic Transcription}
The analysis uses 17 Universal Dependencies (UD v2) treebanks \citep{nivre2020universal} spanning nine language families (training splits for 16 languages and the test split for Yoruba): \texttt{UD\_English-EWT} (\texttt{en}), \texttt{UD\_German-GSD} (\texttt{de}), \texttt{UD\_Portuguese-Bosque} (\texttt{pt}), \texttt{UD\_Spanish-GSD} (\texttt{es}), \texttt{UD\_French-GSD} (\texttt{fr}), \texttt{UD\_Polish-PDB} (\texttt{pl}), \texttt{UD\_Russian-GSD} (\texttt{ru}), \texttt{UD\_Hindi-HDTB} (\texttt{hi}), \texttt{UD\_Hungarian-Szeged} (\texttt{hu}), \texttt{UD\_Finnish-TDT} (\texttt{fi}), \texttt{UD\_Turkish-IMST} (\texttt{tr}), \texttt{UD\_Indonesian-GSD} (\texttt{id}), \texttt{UD\_Korean-GSD} (\texttt{ko}), \texttt{UD\_Tamil-TTB} (\texttt{ta}), \texttt{UD\_Maltese-MUDT} (\texttt{mt}), \texttt{UD\_Vietnamese-VTB} (\texttt{vi}), and \texttt{UD\_Yoruba-YTB} (\texttt{yo}). Non-alphabetic tokens, multiword token ranges, and punctuation (\texttt{UPOS == PUNCT}) were excluded, and remaining word forms were lowercased and counted by token frequency.

All 17 languages were transcribed into IPA using a single unified pipeline---Epitran v1.35.3 \citep{mortensen2018epitran}---achieving 99.7\%--100.0\% token coverage (100.0\% for English). Whereas earlier work on English \citep{araujo2013thesis} relied on CMUdict, here English (\texttt{eng-Latn}) is transcribed through Epitran backed by Flite's \texttt{lex\_lookup}\footnote{Flite (Festival Lite) is a lightweight speech synthesis engine whose \texttt{lex\_lookup} utility queries an English pronunciation lexicon derived from CMUdict and falls back to letter-to-sound rules for out-of-vocabulary words. Unlike the other 16 languages, for which Epitran uses internal orthographic-to-IPA rewrite rules, English (\texttt{eng-Latn}) has an irregular orthography that Epitran handles by delegating to this external binary.} so that every language in the sample is processed on the same methodological footing. IPA strings were segmented using PanPhon's \texttt{ipa\_segs()} parser, which treats tied affricates and diacritic-modified bases recognized by PanPhon as single segments while skipping unsupported suprasegmental/stress symbols. Across all 17 corpora, multi-character IPA units merged as single segments range from $2$ to $33$ segment types ($0.23\%$ of segment tokens in Spanish to $24.83\%$ in German; $3.30\%$ in English), and skipped non-segmental characters range from $0.00\%$ (German, Portuguese) to $8.55\%$ (Finnish; see \texttt{results/major\_vs\_finer\_and\_audit.csv}). Feature vectors were extracted with PanPhon v0.22.2, and all numerical computations used NumPy v2.5.5 and SciPy v1.18.1. Note that because these G2P outputs reflect rule-based or lexicon-backed pronunciation transcriptions rather than language-specific phonemic analyses, we refer to the resulting units as transcribed segments throughout.

\subsection{Distinctive-Feature Representation and Triphone Metrics}
Building on the distinctive-feature framework reviewed in Section~2.2, using articulatory feature vectors rather than atomic phoneme symbols or raw acoustic waveforms is essential for three reasons: (i) atomic symbols treat every pair of distinct segments as equally different, whereas feature vectors yield a graded distance reflecting how many articulatory and perceptual properties two sounds share (for example, /p/ and /b/ differ only in $[\pm\text{voice}]$, whereas /p/ and /a/ differ across major class, manner, and place); (ii) universal features embed segments from typologically unrelated phoneme inventories into a single, directly comparable coordinate space; and (iii) feature representations provide a tractable proxy for articulatory and perceptual similarity across 17 languages without requiring large-scale, force-aligned acoustic corpora.

Each IPA segment $\phi$ was mapped to a 24-dimensional ternary articulatory feature vector $\mathbf{v}(\phi) \in \{-1, 0, +1\}^{24}$ using PanPhon v0.22.2 \citep[extending the 21-feature core of][with \texttt{velaric}, \texttt{hireg}, and \texttt{hitone}/\texttt{long} dimensions]{mortensen2016panphon}, where $+1$ indicates the presence of a feature, $-1$ indicates its absence, and $0$ indicates that the feature is unspecified or non-applicable.\footnote{In PanPhon's uniform cross-linguistic representation, $0$ prevents false articulatory oppositions when a feature is physically non-applicable to a given sound class---for instance, coronal place features such as \texttt{ant} and \texttt{distr} distinguish consonants like [s] and [\textipa{S}], but are non-applicable to vowels or glottal segments, and tonal features (\texttt{hitone}, \texttt{hireg}) are $0$ for non-tonal segments. Forcing such entries to $-1$ would falsely equate ``not applicable'' with actively possessing the opposite gesture. Because segments vary in their number of $0$ entries, comparing both unnormalized Hamming distance and $\ell_2$-normalized cosine distance ensures the results are not artifacts of feature sparsity.} The 24 features span five phonological dimensions:
\begin{itemize}
    \item \emph{Major class (3):} syllabic (\texttt{syl}), sonorant (\texttt{son}), consonantal (\texttt{cons});
    \item \emph{Manner (5):} continuant (\texttt{cont}), delayed release (\texttt{delrel}), lateral (\texttt{lat}), nasal (\texttt{nas}), strident (\texttt{strid});
    \item \emph{Laryngeal (3):} voice (\texttt{voi}), spread glottis (\texttt{sg}), constricted glottis (\texttt{cg});
    \item \emph{Place and vowel quality (8):} anterior (\texttt{ant}), coronal (\texttt{cor}), distributed (\texttt{distr}), labial (\texttt{lab}), high (\texttt{hi}), low (\texttt{lo}), back (\texttt{back}), round (\texttt{round});
    \item \emph{Airstream, length, and tone (5):} velaric (\texttt{velaric}), tense (\texttt{tense}), long (\texttt{long}), high tone (\texttt{hitone}), high register (\texttt{hireg}).
\end{itemize}

To ensure that the findings are not artifacts of a single distance formulation, we evaluate two complementary feature-vector metrics:
\begin{enumerate}
    \item \emph{Feature Hamming distance:} the number of non-shared distinctive features,
    \begin{equation}
    d_{\text{H}}(\phi_a, \phi_b) = \sum_{i=1}^{24} \mathbf{1}_{\{v_i(\phi_a) \neq v_i(\phi_b)\}} \text{.}
    \end{equation}
    \item \emph{Rescaled cosine distance:} normalized by the $\ell_2$ norm of each segment vector to equalize magnitudes across segments with different numbers of unspecified ($0$) entries,
    \begin{equation}
    d_{\cos}(\phi_a, \phi_b) = \frac{1}{2}\left(1 - \frac{\mathbf{v}(\phi_a) \cdot \mathbf{v}(\phi_b)}{\lVert\mathbf{v}(\phi_a)\rVert_2 \, \lVert\mathbf{v}(\phi_b)\rVert_2}\right) \in [0, 1] \text{.}
    \end{equation}
    For joint entropy and modal peak estimation under $d_{\cos}$, the interval $[0, 1]$ is discretized into $20$ uniform bins of width $0.05$ (whereas $\mathbb{E}[d_{\cos}]$, $\sigma(d_{\cos})$, and $r_{\cos}$ are computed directly from the unbinned continuous values).
\end{enumerate}

For every word of phonemic length $L \geq 3$ with segment sequence $(\phi_1, \dots, \phi_L)$ and corpus token count $w$, adjacent distances $d_i = d(\phi_i, \phi_{i+1})$ are computed for $i=1,\dots,L-1$, and all contiguous triphones $(\phi_i, \phi_{i+1}, \phi_{i+2})$ ($i=1,\dots,L-2$) with intradistance pair $(d_i, d_{i+1})$ are extracted and weighted by $w$. Note that the overall expected value $\mathbb{E}[d]$ and standard deviation $\sigma(d)$ in Tables~\ref{tab:comparison}--\ref{tab:cosine} are computed across all $L-1$ adjacent transitions per word, whereas $\mathbb{E}[d_1]$ and $\mathbb{E}[d_2]$ in Table~\ref{tab:corpora} average over the $L-2$ triphone windows (so $d_1$ excludes the word-final transition $d_{L-1}$ and $d_2$ excludes the word-initial transition $d_1$). From the token-weighted empirical distribution $P(d_1, d_2)$, the following metrics are computed:
\begin{itemize}
    \item The modal intradistance pair $(d_1^*, d_2^*) = \arg\max_{(d_1,d_2) \in \mathcal{D}} P(d_1, d_2)$, where $\mathcal{D} = \{0,\dots,24\}^2$ for $d_{\text{H}}$ (or the $20 \times 20$ bin grid for $d_{\cos}$);
    \item The expected value $\mathbb{E}[d]$ and standard deviation $\sigma(d)$ of adjacent transitions;
    \item The joint Shannon entropy $H(d_1, d_2) = -\sum_{d_1,d_2} P(d_1, d_2)\log_2 P(d_1, d_2)$ (in bits);
    \item The within-word Pearson correlation $r$ between the mean and standard deviation of $\{d_1,\dots,d_{L-1}\}$;
    \item The middle-phone directional asymmetry, comparing $\mathbb{E}[d_1 \mid \phi_2]$ against $\mathbb{E}[d_2 \mid \phi_2]$ across middle segments $\phi_2$ with token weight $\geq 50$ via a paired Wilcoxon signed-rank test (where $W = \min(W^+, W^-)$ denotes the minimum of the positive and negative signed-rank sums).
\end{itemize}

\subsection{Nested Permutation Baselines and Inferential Testing}
To separate genuine phonotactic ordering from marginal phoneme frequencies, word-length effects, and basic consonant--vowel (C/V) alternation, we construct two nested permutation baselines:
\begin{enumerate}
    \item \emph{Unrestricted within-word shuffle:} for each word type, its segment sequence $(\phi_1,\dots,\phi_L)$ is randomly permuted while preserving its exact segment multiset and token count $w$. Tables~\ref{tab:comparison}--\ref{tab:cosine} report a fixed-seed permutation (\texttt{seed=42}), and 20 independent Monte Carlo seeds (\texttt{1000--1019}) are run per language to quantify shuffle variance.
    \item \emph{Within-word C/V-template-preserving shuffle:} each segment is classified by PanPhon's syllabic feature as syllabic (\texttt{syl} $= +1$, capturing vowels and syllabic sonorants such as English [\textipa{\*r}\textsubring{}]) or non-syllabic (\texttt{syl} $\neq +1$, capturing consonants and glides). Every word's exact C/V template and exact segment multiset are both held fixed while permuting C segments across its C positions and V segments across its V positions within each word ($20$ Monte Carlo seeds, \texttt{2000--2019}). To pinpoint what drives residual joint-entropy differences beyond C/V alternation, we decompose joint entropy as $H(d_1, d_2) = H(d_1) + H(d_2) - I(d_1; d_2)$, separating marginal entropy reduction ($H(d_1)+H(d_2)$) from sequential coupling between incoming and outgoing transitions (mutual information $I(d_1; d_2)$).
\end{enumerate}
Cross-linguistic consistency across the $n=17$ languages is evaluated using two-sided paired Wilcoxon signed-rank tests and exact binomial sign tests.

\section{Results}
\begin{figure}[htbp]
\centering
\includegraphics[width=\textwidth]{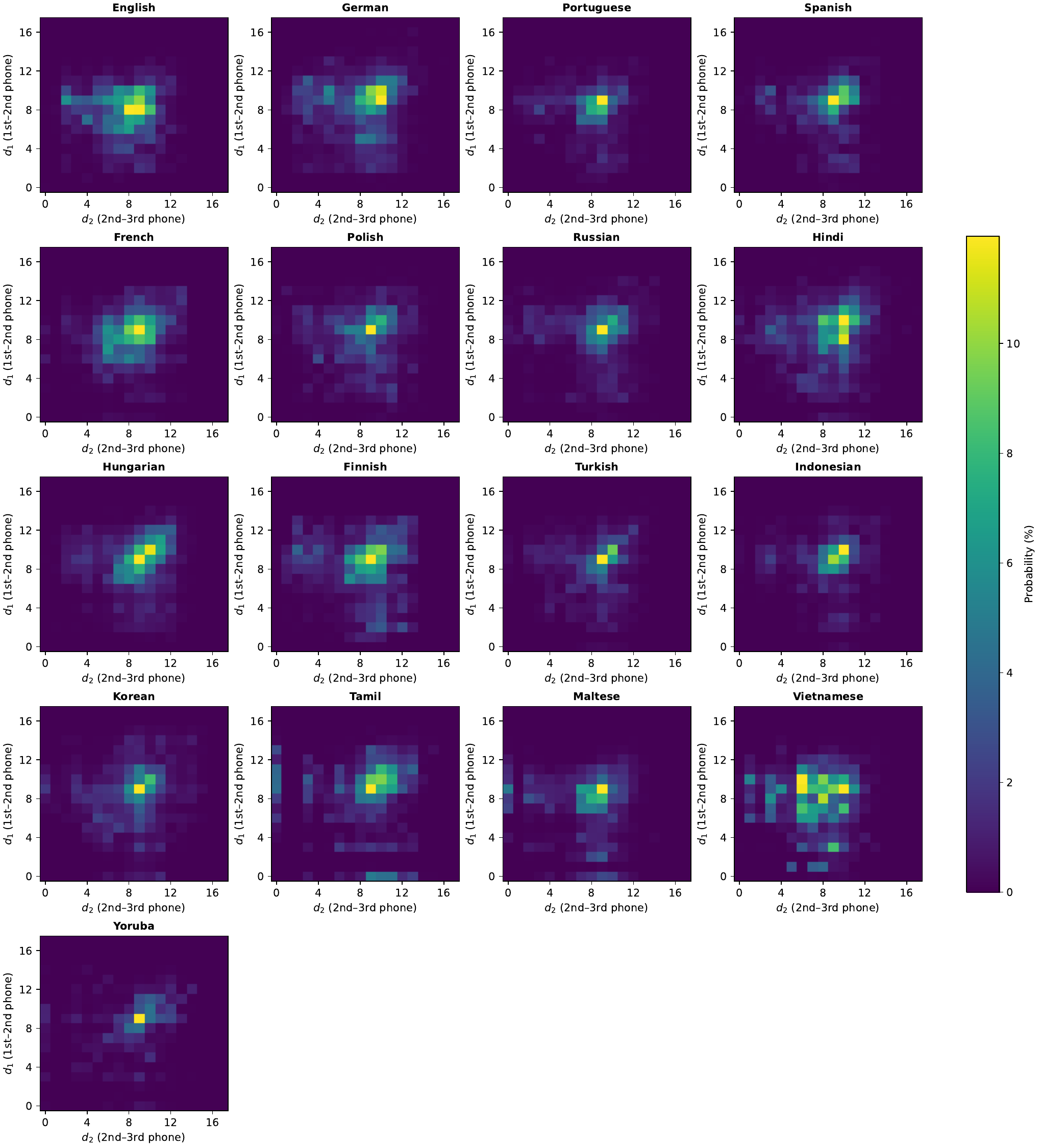}
\caption{Empirical joint probability distributions $P(d_1, d_2)$ of adjacent distinctive-feature distances in segmental triphones under feature Hamming distance $d_{\text{H}}$ across 17 languages (axes shown over $0$--$17$ features).}
\label{fig:heatmaps}
\end{figure}

Table~\ref{tab:corpora} summarizes the corpus statistics and triphone peaks across the 17 languages, and Figure~\ref{fig:heatmaps} displays the empirical joint distributions $P(d_1, d_2)$. In 16 of the 17 languages, $P(d_1, d_2)$ concentrates around an intermediate-distance peak at $(d_1^*, d_2^*) \in \{(8,8), (9,9), (9,10), (10,10)\}$ (with Vietnamese peaking at $(10,6)$), with very low mass at small ($d \leq 3$) or large ($d \geq 14$) feature distances.

\begin{table}[htbp]
\tbl{Corpus statistics, effective segment inventory size $|P|$, \% CVC/VCV alternation, modal triphone intradistance peak $(d_1^*, d_2^*)$, and directional asymmetry under feature Hamming distance $d_{\text{H}}$ across 17 languages.}
{\begin{tabular}{llrrrrrcc}
\toprule
Language & Family & Tokens & $|P|$ & CVC/VCV & Triphones & Peak $(d_1^*, d_2^*)$ & $\mathbb{E}[d_1], \mathbb{E}[d_2]$ & Mid $d_1>d_2$ ($p$) \\
\midrule
English (\texttt{en})    & IE (Germanic)    & 176,883 & 40 & 59.2\% & 351,106 & $(8,8)$ (4.7\%)   & 7.80, 7.44 & 50.0\% (0.98) \\
German (\texttt{de})     & IE (Germanic)    & 222,741 & 59 & 52.9\% & 718,194 & $(9,10)$ (4.9\%)  & 8.31, 7.99 & 48.0\% (0.72) \\
Portuguese (\texttt{pt}) & IE (Romance)     & 146,957 & 41 & 63.5\% & 396,625 & $(9,9)$ (9.7\%)   & 8.00, 7.93 & 45.5\% (0.76) \\
Spanish (\texttt{es})    & IE (Romance)     & 331,365 & 44 & 61.2\% & 916,794 & $(9,9)$ (7.9\%)   & 8.35, 8.43 & 50.0\% (0.62) \\
French (\texttt{fr})     & IE (Romance)     & 307,316 & 51 & 67.5\% & 571,087 & $(9,9)$ (5.1\%)   & 8.38, 8.17 & 62.9\% (0.29) \\
Polish (\texttt{pl})     & IE (Slavic)      & 232,578 & 41 & 59.6\% & 841,156 & $(9,9)$ (6.6\%)   & 8.25, 8.30 & 43.2\% (0.89) \\
Russian (\texttt{ru})    & IE (Slavic)      &  58,306 & 54 & 60.0\% & 246,454 & $(9,9)$ (7.8\%)   & 8.57, 8.38 & 54.8\% (0.99) \\
Hindi (\texttt{hi})      & IE (Indo-Aryan)  & 259,658 & 64 & 65.5\% & 589,202 & $(10,10)$ (4.8\%) & 8.52, 8.44 & 40.4\% (0.03) \\
Hungarian (\texttt{hu})  & Uralic           &  17,043 & 63 & 69.7\% &  66,747 & $(9,9)$ (5.7\%)   & 8.77, 8.76 & 56.5\% (0.44) \\
Finnish (\texttt{fi})    & Uralic           & 135,519 & 37 & 61.0\% & 660,825 & $(9,9)$ (4.8\%)   & 8.27, 8.33 & 64.5\% (0.33) \\
Turkish (\texttt{tr})    & Turkic           &  30,973 & 32 & 70.3\% & 127,704 & $(9,9)$ (9.4\%)   & 8.34, 8.39 & 44.8\% (0.98) \\
Indonesian (\texttt{id}) & Austronesian     &  80,944 & 41 & 73.0\% & 303,829 & $(10,10)$ (9.0\%) & 8.73, 8.65 & 40.7\% (0.41) \\
Korean (\texttt{ko})     & Koreanic         &  49,011 & 37 & 62.3\% & 241,967 & $(9,9)$ (6.7\%)   & 8.38, 8.28 & 58.6\% (0.92) \\
Tamil (\texttt{ta})      & Dravidian        &   5,562 & 32 & 68.6\% &  28,993 & $(9,9)$ (5.1\%)   & 8.44, 8.35 & 39.3\% (0.81) \\
Maltese (\texttt{mt})    & Afro-Asiatic     &  20,056 & 52 & 51.0\% &  55,041 & $(9,9)$ (6.9\%)   & 7.37, 7.29 & 48.5\% (0.67) \\
Vietnamese (\texttt{vi}) & Austroasiatic    &  17,028 & 35 & 51.6\% &  28,694 & $(10,6)$ (3.7\%)  & 7.78, 7.21 & 50.0\% (0.93) \\
Yoruba (\texttt{yo})     & Niger-Congo      &   7,102 & 32 & 85.7\% &   8,532 & $(9,9)$ (12.0\%)  & 8.57, 8.54 & 28.0\% (0.10) \\
\bottomrule
\end{tabular}}
\label{tab:corpora}
\end{table}

\begin{figure}[htbp]
\centering
\includegraphics[width=\textwidth]{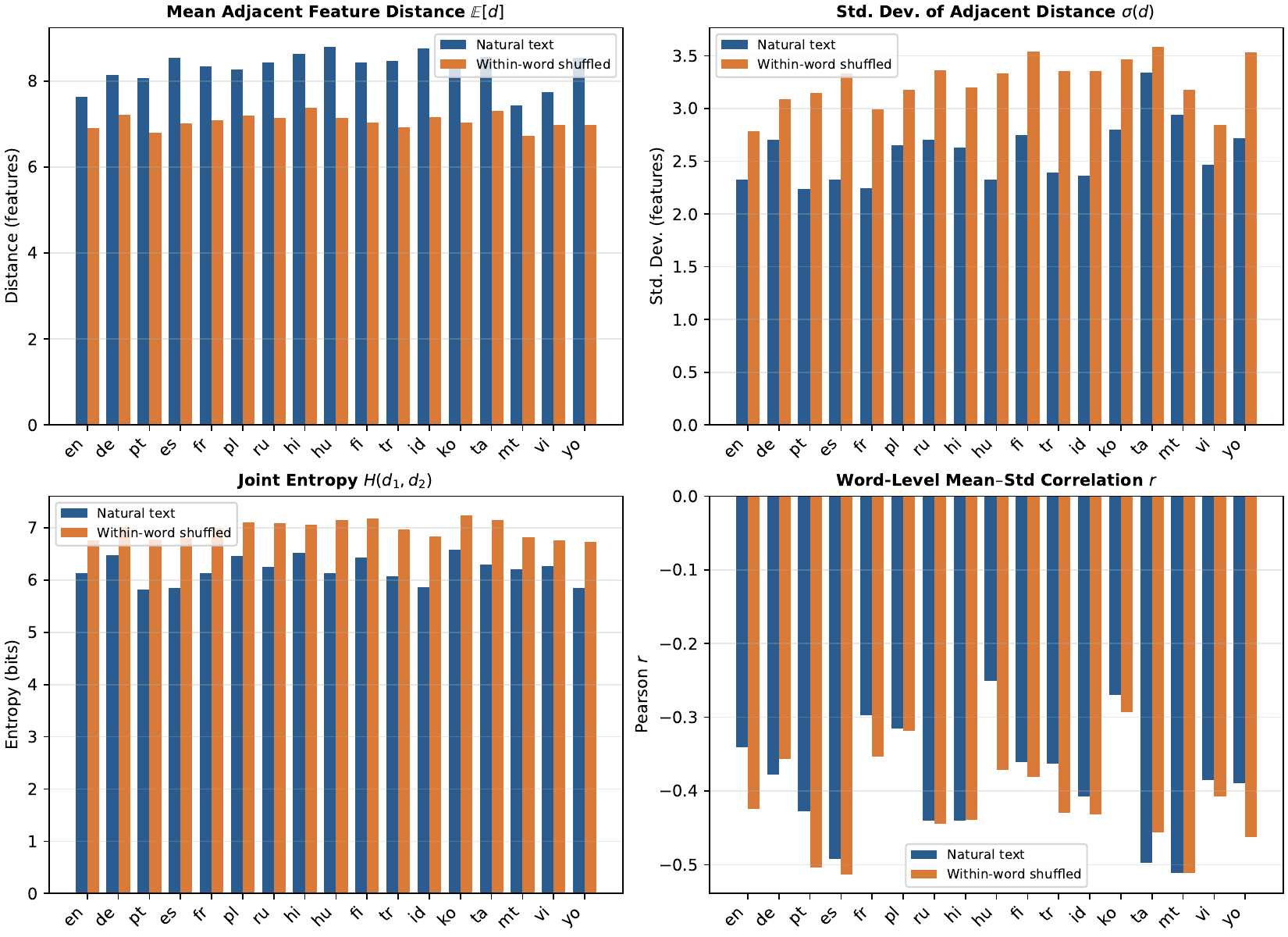}
\caption{Natural text vs.\ within-word shuffled baseline across 17 languages under feature Hamming distance $d_{\text{H}}$: expected adjacent distance $\mathbb{E}[d]$, standard deviation $\sigma(d)$, joint triphone distance entropy $H(d_1, d_2)$, and word-level mean--std correlation $r$.}
\label{fig:comparison}
\end{figure}

Comparing natural text against the within-word shuffled baseline (Table~\ref{tab:comparison} and Figure~\ref{fig:comparison}) rejects the inventory-only null hypothesis $H_{0,\text{unrest}}$ for all three metrics across the 17 languages (two-sided exact binomial sign test and paired Wilcoxon signed-rank test across $n=17$ languages: $W=0$, $p = 1.53 \times 10^{-5}$):
\begin{enumerate}
    \item \emph{Upward shift in adjacent contrast:} Natural phonotactics increases the expected adjacent feature distance ($\mathbb{E}[d]_{\text{real}} > \mathbb{E}[d]_{\text{shuf}}$) in every language, by $+0.71$ (Maltese) to $+1.66$ (Hungarian) features.
    \item \emph{Narrower distance dispersion:} Natural words have consistently lower standard deviation $\sigma(d)$ than their shuffled counterparts, reduced by $0.23$ to $1.01$ features (e.g., $2.24$ vs.\ $3.14$ in Portuguese; $2.33$ vs.\ $3.33$ in Hungarian).
    \item \emph{Reduced joint entropy:} The joint entropy $H(d_1, d_2)$ drops by $0.50$ to $1.01$ bits in natural text relative to shuffled words, demonstrating that phonotactic sequencing concentrates transitions into the intermediate-distance band.
\end{enumerate}
Across 20 independent Monte Carlo shuffle seeds per language, the shuffle standard deviations remain small ($\text{SD}_{\text{shuf}} \in [0.015, 0.113]$ for $\mathbb{E}[d]$, $[0.007, 0.044]$ for $\sigma(d)$, and $[0.005, 0.046]$ bits for $H(d_1, d_2)$), confirming that every language's real-versus-shuffled gap far exceeds permutation noise ($W=0$, $p = 1.53 \times 10^{-5}$ against the 20-seed shuffle means). Moreover, all three $17/17$ directional contrasts hold identically when subsampling every larger corpus down to $17{,}000$ tokens (\texttt{seed=123}) and when weighting triphones by unique word types rather than corpus tokens.

By contrast, two patterns from single-language analyses do not hold as cross-linguistic invariants under baseline controls. First, while the within-word correlation $r$ between mean distance and standard deviation is negative in all 17 languages ($r \in [-0.511, -0.251]$), virtually identical negative correlations appear in the shuffled baseline ($r_{\text{shuf}} \in [-0.512, -0.280]$), indicating a combinatorial property of finite bounded sequences rather than a uniquely phonotactic constraint. Second, the directional asymmetry $d_1 > d_2$ observed in English ($\mathbb{E}[d_1]=7.80 > \mathbb{E}[d_2]=7.44$, $p=0.98$) is absent or reversed in several languages (e.g., Spanish, Polish, Turkish, Finnish, Tamil, Yoruba). For Hindi, although the token-weighted mean is slightly higher for $d_1$ ($\mathbb{E}[d_1]=8.52 > \mathbb{E}[d_2]=8.44$), only $40.4\%$ of middle phones exhibit $d_1 > d_2$ (uncorrected $p=0.033$); after Holm--Bonferroni correction across the 17 languages ($\alpha/17 = 0.0029$), none of the 17 languages show statistically significant per-middle-phone directional asymmetry.

\begin{table}[htbp]
\tbl{Comparison of natural text against the within-word shuffled baseline across 17 languages under feature Hamming distance $d_{\text{H}}$.}
{\begin{tabular}{lcccccccc}
\toprule
Language & \multicolumn{2}{c}{$\mathbb{E}[d]$} & \multicolumn{2}{c}{$\sigma(d)$} & \multicolumn{2}{c}{$H(d_1, d_2)$ (bits)} & \multicolumn{2}{c}{Word $r$} \\
\cmidrule(lr){2-3} \cmidrule(lr){4-5} \cmidrule(lr){6-7} \cmidrule(lr){8-9}
 & Real & Shuf. & Real & Shuf. & Real & Shuf. & Real & Shuf. \\
\midrule
English    & \textbf{7.63} & 6.90 & \textbf{2.33} & 2.79 & \textbf{6.14} & 6.77 & $-0.341$ & $-0.425$ \\
German     & \textbf{8.15} & 7.23 & \textbf{2.70} & 3.10 & \textbf{6.47} & 7.02 & $-0.377$ & $-0.356$ \\
Portuguese & \textbf{8.06} & 6.81 & \textbf{2.24} & 3.14 & \textbf{5.82} & 6.79 & $-0.428$ & $-0.512$ \\
Spanish    & \textbf{8.54} & 7.10 & \textbf{2.33} & 3.30 & \textbf{5.85} & 6.78 & $-0.492$ & $-0.508$ \\
French     & \textbf{8.33} & 7.00 & \textbf{2.25} & 2.99 & \textbf{6.14} & 7.01 & $-0.298$ & $-0.360$ \\
Polish     & \textbf{8.27} & 7.20 & \textbf{2.65} & 3.18 & \textbf{6.47} & 7.10 & $-0.315$ & $-0.321$ \\
Russian    & \textbf{8.43} & 7.07 & \textbf{2.71} & 3.39 & \textbf{6.26} & 7.10 & $-0.440$ & $-0.423$ \\
Hindi      & \textbf{8.63} & 7.48 & \textbf{2.63} & 3.17 & \textbf{6.52} & 7.04 & $-0.441$ & $-0.437$ \\
Hungarian  & \textbf{8.80} & 7.17 & \textbf{2.33} & 3.31 & \textbf{6.13} & 7.13 & $-0.251$ & $-0.390$ \\
Finnish    & \textbf{8.43} & 7.01 & \textbf{2.75} & 3.54 & \textbf{6.43} & 7.16 & $-0.361$ & $-0.385$ \\
Turkish    & \textbf{8.47} & 6.90 & \textbf{2.39} & 3.37 & \textbf{6.08} & 6.97 & $-0.363$ & $-0.432$ \\
Indonesian & \textbf{8.75} & 7.20 & \textbf{2.36} & 3.33 & \textbf{5.87} & 6.82 & $-0.407$ & $-0.418$ \\
Korean     & \textbf{8.46} & 7.05 & \textbf{2.80} & 3.48 & \textbf{6.58} & 7.26 & $-0.270$ & $-0.280$ \\
Tamil      & \textbf{8.56} & 7.42 & \textbf{3.34} & 3.58 & \textbf{6.30} & 7.13 & $-0.498$ & $-0.442$ \\
Maltese    & \textbf{7.43} & 6.79 & \textbf{2.94} & 3.12 & \textbf{6.21} & 6.80 & $-0.511$ & $-0.494$ \\
Vietnamese & \textbf{7.74} & 6.98 & \textbf{2.47} & 2.85 & \textbf{6.27} & 6.80 & $-0.385$ & $-0.426$ \\
Yoruba     & \textbf{8.54} & 6.81 & \textbf{2.72} & 3.63 & \textbf{5.85} & 6.78 & $-0.390$ & $-0.431$ \\
\bottomrule
\end{tabular}}
\label{tab:comparison}
\end{table}

\subsection{Robustness Under Cosine Distance}
\begin{figure}[htbp]
\centering
\includegraphics[width=\textwidth]{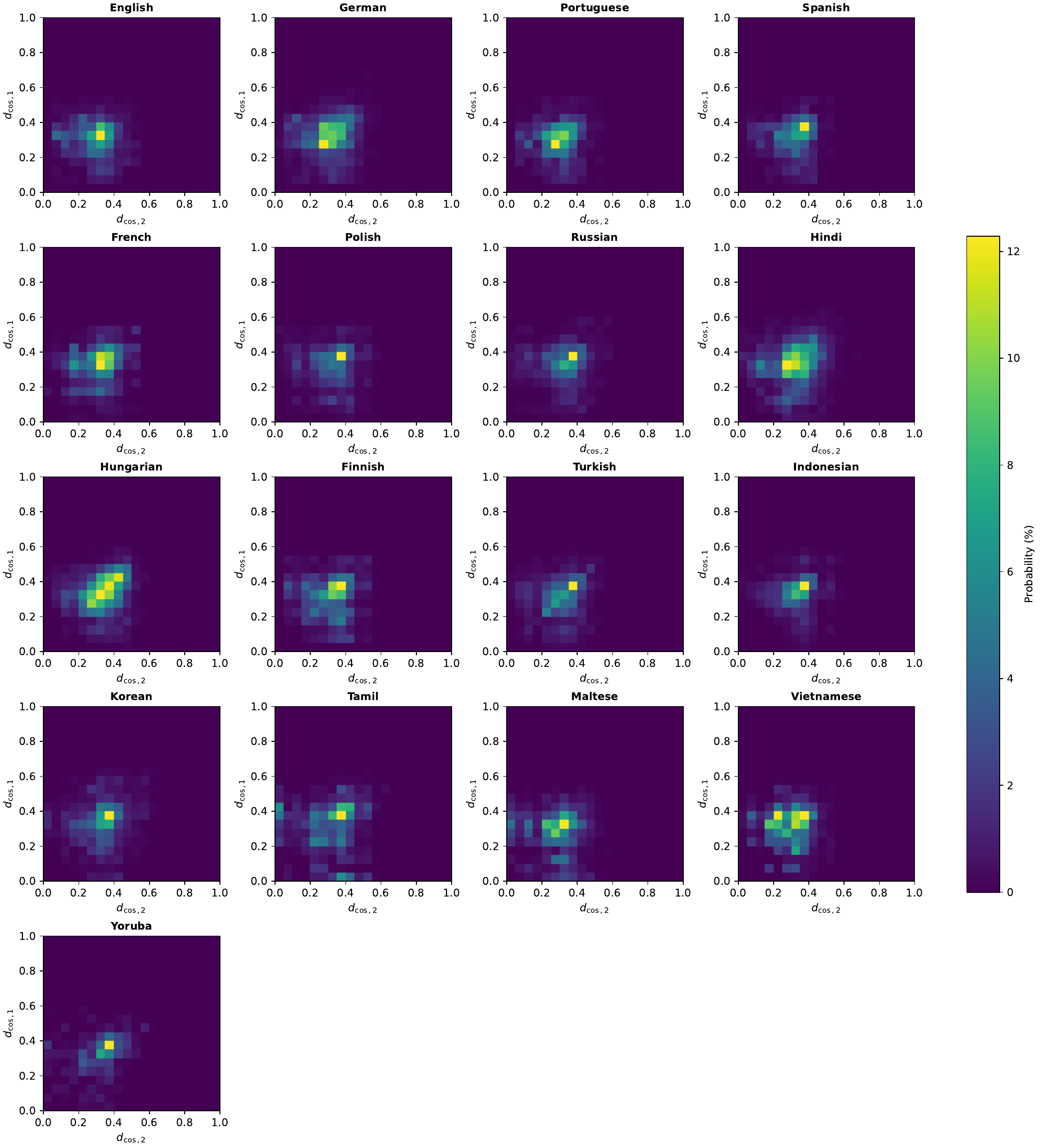}
\caption{Empirical joint probability distributions $P(d_{\cos,1}, d_{\cos,2})$ of adjacent distinctive-feature distances in segmental triphones under rescaled cosine distance $d_{\cos} \in [0, 1]$ across 17 languages.}
\label{fig:heatmaps_cosine}
\end{figure}

\begin{figure}[htbp]
\centering
\includegraphics[width=\textwidth]{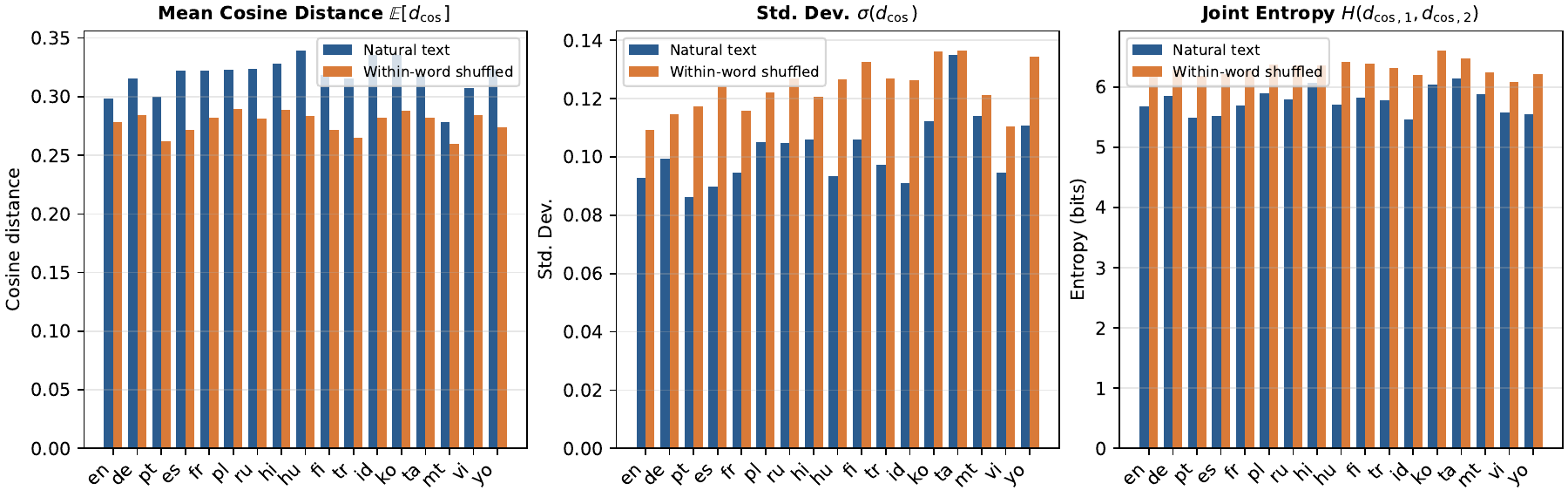}
\caption{Natural text vs.\ within-word shuffled baseline across 17 languages under rescaled cosine distance $d_{\cos}$: expected distance $\mathbb{E}[d_{\cos}]$, standard deviation $\sigma(d_{\cos})$, and joint triphone distance entropy $H(d_{\cos,1}, d_{\cos,2})$.}
\label{fig:cosine}
\end{figure}

To verify that the observed phonotactic dispersion pattern is not driven by shared unspecified ($0$) entries in sparse consonant feature vectors, we repeated the full 17-language analysis using the $\ell_2$-normalized rescaled cosine distance $d_{\cos} \in [0, 1]$. As shown in Table~\ref{tab:cosine} and Figures~\ref{fig:heatmaps_cosine}--\ref{fig:cosine}, every finding replicates across all 17 languages: natural text exhibits an intermediate-distance peak at $(d_{\cos,1}^*, d_{\cos,2}^*) \in [0.28, 0.38]$, a higher expected contrast ($\mathbb{E}[d_{\cos}]_{\text{real}} > \mathbb{E}[d_{\cos}]_{\text{shuf}}$ by $+0.019$ to $+0.055$ in $17/17$ languages), a narrower dispersion ($\sigma(d_{\cos})$ reduced by $0.002$ to $0.035$ in $17/17$ languages), and a reduced joint entropy ($H_{\text{real}} < H_{\text{shuf}}$ by $0.29$ to $0.74$ bits in $17/17$ languages). Moreover, under $d_{\cos}$, the magnitude of the within-word correlation $|r|$ is smaller in natural text than in the shuffled baseline in 16 of 17 languages (Yoruba is $-0.351$ vs.\ $-0.341$), further confirming that negative $r$ is largely a combinatorial baseline artifact rather than a phonotactic law.

\begin{table}[htbp]
\tbl{Robustness check under rescaled cosine distance $d_{\cos} \in [0,1]$: natural text vs.\ within-word shuffled baseline across 17 languages.}
{\begin{tabular}{lccccccccc}
\toprule
Language & Peak $(d_{\cos,1}^*, d_{\cos,2}^*)$ & \multicolumn{2}{c}{$\mathbb{E}[d_{\cos}]$} & \multicolumn{2}{c}{$\sigma(d_{\cos})$} & \multicolumn{2}{c}{$H(d_{\cos,1}, d_{\cos,2})$} & \multicolumn{2}{c}{Word $r_{\cos}$} \\
\cmidrule(lr){3-4} \cmidrule(lr){5-6} \cmidrule(lr){7-8} \cmidrule(lr){9-10}
 & & Real & Shuf. & Real & Shuf. & Real & Shuf. & Real & Shuf. \\
\midrule
English    & $(0.33, 0.33)$ & \textbf{0.299} & 0.278 & \textbf{0.093} & 0.109 & \textbf{5.67} & 6.17 & $-0.269$ & $-0.384$ \\
German     & $(0.28, 0.28)$ & \textbf{0.315} & 0.286 & \textbf{0.099} & 0.115 & \textbf{5.85} & 6.28 & $-0.186$ & $-0.263$ \\
Portuguese & $(0.28, 0.28)$ & \textbf{0.300} & 0.263 & \textbf{0.086} & 0.117 & \textbf{5.49} & 6.20 & $-0.231$ & $-0.379$ \\
Spanish    & $(0.38, 0.38)$ & \textbf{0.322} & 0.274 & \textbf{0.090} & 0.123 & \textbf{5.52} & 6.18 & $-0.355$ & $-0.414$ \\
French     & $(0.33, 0.33)$ & \textbf{0.322} & 0.279 & \textbf{0.094} & 0.116 & \textbf{5.69} & 6.26 & $-0.324$ & $-0.338$ \\
Polish     & $(0.38, 0.38)$ & \textbf{0.323} & 0.289 & \textbf{0.105} & 0.123 & \textbf{5.89} & 6.38 & $-0.227$ & $-0.251$ \\
Russian    & $(0.38, 0.38)$ & \textbf{0.324} & 0.279 & \textbf{0.105} & 0.128 & \textbf{5.79} & 6.37 & $-0.296$ & $-0.332$ \\
Hindi      & $(0.33, 0.28)$ & \textbf{0.328} & 0.292 & \textbf{0.106} & 0.120 & \textbf{6.07} & 6.39 & $-0.251$ & $-0.302$ \\
Hungarian  & $(0.33, 0.33)$ & \textbf{0.339} & 0.285 & \textbf{0.093} & 0.126 & \textbf{5.71} & 6.41 & $-0.089$ & $-0.291$ \\
Finnish    & $(0.38, 0.38)$ & \textbf{0.318} & 0.271 & \textbf{0.106} & 0.132 & \textbf{5.83} & 6.39 & $-0.204$ & $-0.280$ \\
Turkish    & $(0.38, 0.38)$ & \textbf{0.315} & 0.264 & \textbf{0.097} & 0.127 & \textbf{5.78} & 6.31 & $-0.242$ & $-0.329$ \\
Indonesian & $(0.38, 0.38)$ & \textbf{0.335} & 0.283 & \textbf{0.091} & 0.126 & \textbf{5.46} & 6.18 & $-0.263$ & $-0.348$ \\
Korean     & $(0.38, 0.38)$ & \textbf{0.336} & 0.288 & \textbf{0.112} & 0.137 & \textbf{6.04} & 6.62 & $-0.176$ & $-0.224$ \\
Tamil      & $(0.38, 0.38)$ & \textbf{0.318} & 0.286 & \textbf{0.135} & 0.136 & \textbf{6.14} & 6.45 & $-0.300$ & $-0.310$ \\
Maltese    & $(0.33, 0.33)$ & \textbf{0.278} & 0.263 & \textbf{0.114} & 0.120 & \textbf{5.89} & 6.22 & $-0.359$ & $-0.366$ \\
Vietnamese & $(0.38, 0.38)$ & \textbf{0.307} & 0.284 & \textbf{0.095} & 0.111 & \textbf{5.58} & 6.08 & $-0.309$ & $-0.413$ \\
Yoruba     & $(0.38, 0.38)$ & \textbf{0.324} & 0.266 & \textbf{0.111} & 0.138 & \textbf{5.56} & 6.24 & $-0.351$ & $-0.341$ \\
\bottomrule
\end{tabular}}
\label{tab:cosine}
\end{table}

\subsection{Separating C/V Alternation from Sub-Syllabic Sequencing via Mutual Information}
To test the C/V-template-only null hypothesis $H_{0,\text{CV}}$ under the strict within-word C/V-template-preserving shuffle (20 Monte Carlo seeds), Table~\ref{tab:cv_decomp} compares natural text against the 20-seed C/V-preserving means and decomposes joint entropy into marginal entropies and mutual information, $H(d_1, d_2) = H(d_1) + H(d_2) - I(d_1; d_2)$:
\begin{itemize}
    \item \emph{Expected contrast $\mathbb{E}[d]$ and dispersion $\sigma(d)$:} Once each word's segment multiset and C/V template are both held fixed, neither $\mathbb{E}[d]$ ($4/17$ languages higher, $W = 35.0$, $p = 0.051$) nor $\sigma(d)$ ($11/17$ languages lower, $W = 62.0$, $p = 0.517$) shows a statistically significant separation (noting that non-significance indicates absence of evidence for a systematic directional shift rather than proof of equivalence).
    \item \emph{Joint entropy $H(d_1, d_2)$ and mutual information $I(d_1; d_2)$:} Natural sequences exhibit significantly lower joint entropy $H(d_1, d_2)$ in $16/17$ languages ($W = 4.0$, $p = 1.07 \times 10^{-4}$; Hindi is the sole exception at $6.523$ vs.\ $6.475$). Decomposing $H(d_1, d_2)$ shows that this arises from both lower marginal entropy $H(d_1)+H(d_2)$ ($13/17$ languages, $W = 24.0$, $p = 0.011$) and, most consistently, \textbf{higher mutual information $I(d_1; d_2)$ in $17/17$ languages} ($+0.007$ to $+0.101$ bits, $W = 0.0$, $p = 1.53 \times 10^{-5}$; $9/9$ language families). Thus, even within fixed C/V templates and fixed word-level segment multisets, incoming transition distance $d_1$ constrains outgoing transition distance $d_2$ more strongly in natural speech than in randomized C/V-preserving sequences.
\end{itemize}

\begin{table}[htbp]
\tbl{Within-word C/V-template-preserving baseline (20 Monte Carlo seeds, mean $\pm$ SD) and mutual-information decomposition $H(d_1, d_2) = H(d_1) + H(d_2) - I(d_1; d_2)$ across 17 languages (Hamming distance $d_{\text{H}}$).}
{\begin{tabular}{lccccccc}
\toprule
Language & \multicolumn{2}{c}{$H(d_1)+H(d_2)$} & \multicolumn{2}{c}{$H(d_1, d_2)$ (bits)} & \multicolumn{2}{c}{$I(d_1; d_2)$ (bits)} \\
\cmidrule(lr){2-3} \cmidrule(lr){4-5} \cmidrule(lr){6-7}
 & Real & C/V (20 seeds) & Real & C/V (20 seeds) & Real & C/V (20 seeds) \\
\midrule
English    & 6.293 & 6.377 & \textbf{6.138} & $6.288 \pm 0.013$ & \textbf{0.154} & $0.090 \pm 0.005$ \\
German     & 6.611 & 6.721 & \textbf{6.474} & $6.619 \pm 0.011$ & \textbf{0.138} & $0.103 \pm 0.004$ \\
Portuguese & 6.020 & 6.041 & \textbf{5.823} & $5.915 \pm 0.008$ & \textbf{0.197} & $0.125 \pm 0.004$ \\
Spanish    & 6.057 & 6.150 & \textbf{5.846} & $6.021 \pm 0.009$ & \textbf{0.211} & $0.128 \pm 0.004$ \\
French     & 6.281 & 6.447 & \textbf{6.139} & $6.351 \pm 0.009$ & \textbf{0.142} & $0.096 \pm 0.003$ \\
Polish     & 6.658 & 6.644 & \textbf{6.467} & $6.551 \pm 0.009$ & \textbf{0.192} & $0.093 \pm 0.002$ \\
Russian    & 6.430 & 6.536 & \textbf{6.256} & $6.439 \pm 0.006$ & \textbf{0.174} & $0.097 \pm 0.002$ \\
Hindi      & 6.704 & 6.605 & 6.523 & $\mathbf{6.475 \pm 0.023}$ & \textbf{0.181} & $0.130 \pm 0.006$ \\
Hungarian  & 6.309 & 6.364 & \textbf{6.135} & $6.210 \pm 0.008$ & \textbf{0.174} & $0.154 \pm 0.004$ \\
Finnish    & 6.617 & 6.701 & \textbf{6.431} & $6.571 \pm 0.004$ & \textbf{0.186} & $0.130 \pm 0.002$ \\
Turkish    & 6.348 & 6.282 & \textbf{6.075} & $6.089 \pm 0.009$ & \textbf{0.273} & $0.193 \pm 0.003$ \\
Indonesian & 6.025 & 6.057 & \textbf{5.871} & $5.910 \pm 0.024$ & \textbf{0.155} & $0.147 \pm 0.006$ \\
Korean     & 6.780 & 6.822 & \textbf{6.582} & $6.725 \pm 0.006$ & \textbf{0.198} & $0.096 \pm 0.002$ \\
Tamil      & 6.499 & 6.521 & \textbf{6.302} & $6.390 \pm 0.022$ & \textbf{0.197} & $0.131 \pm 0.008$ \\
Maltese    & 6.385 & 6.466 & \textbf{6.213} & $6.352 \pm 0.014$ & \textbf{0.173} & $0.114 \pm 0.004$ \\
Vietnamese & 6.480 & 6.666 & \textbf{6.265} & $6.546 \pm 0.016$ & \textbf{0.215} & $0.120 \pm 0.008$ \\
Yoruba     & 6.313 & 6.297 & \textbf{5.846} & $5.883 \pm 0.034$ & \textbf{0.467} & $0.414 \pm 0.021$ \\
\bottomrule
\end{tabular}}
\label{tab:cv_decomp}
\end{table}

\subsection{Sensitivity Analyses: Feature Weighting, Sub-Syllabic Ablation, Binning, and Family Balance}
To verify that these findings do not depend on equal feature weighting, the treatment of unspecified ($0$) entries, discretization resolution, or Indo-European representation ($8/17$ corpora), we ran four additional checks:
\begin{enumerate}
    \item \emph{PanPhon salience-weighted and unspecified-aware metrics:} Using PanPhon's built-in salience-weighted feature distance and partial Hamming distance (which penalizes $0$-versus-$\pm 1$ mismatches at half the weight of $+1$-versus-$-1$ oppositions), both metrics reject $H_{0,\text{unrest}}$ across $17/17$ languages for $\mathbb{E}[d]$, $\sigma(d)$, and $H(d_1, d_2)$ ($W=0$, $p = 1.53 \times 10^{-5}$). Against the C/V-template-preserving baseline ($H_{0,\text{CV}}$), $\mathbb{E}[d]$ and $\sigma(d)$ show no consistent residual difference ($p \ge 0.071$), whereas natural sequences remain significantly lower in joint entropy $H(d_1, d_2)$ across $17/17$ languages under both weighted ($W=0$, $p = 1.53 \times 10^{-5}$) and partial Hamming ($W=0$, $p = 1.53 \times 10^{-5}$) metrics.
    \item \emph{Major-class vs.\ finer-feature decomposition and non-major-class ablation:} Decomposing the expected adjacent Hamming distance $\mathbb{E}[d]$ into major-class features (\texttt{syl}, \texttt{son}, \texttt{cons}) versus the remaining 21 finer features shows that major-class shifts contribute $23.5\%$--$27.3\%$ ($1.94$--$2.27$ features) across the 17 languages, whereas finer manner, laryngeal, and place/vowel contrasts account for $72.7\%$--$76.5\%$ ($5.49$--$6.56$ features). Restricting the Hamming metric exclusively to the 16 non-major-class articulatory features (manner, laryngeal, and place/vowel dimensions, completely removing \texttt{syl}, \texttt{son}, and \texttt{cons}) preserves all three $17/17$ contrasts against $H_{0,\text{unrest}}$ ($p = 1.53 \times 10^{-5}$). Against $H_{0,\text{CV}}$, natural sequences still exhibit significantly lower joint entropy in $14/17$ languages ($p = 3.85 \times 10^{-3}$; exceptions are Hindi, Tamil, and Maltese), confirming that finer place, manner, and laryngeal transitions contribute to the sub-syllabic entropy reduction beyond C/V alternation.
    \item \emph{Cosine discretization and multiple-testing control:} Varying the cosine histogram resolution to $10$ or $40$ bins yields $H_{\text{real}} < H_{\text{shuf}}$ in $17/17$ languages under both settings. Applying Holm--Bonferroni correction across the six primary Wilcoxon tests ($3$ metrics $\times$ $2$ baselines) leaves all four rejected hypotheses significant (adjusted $p = 9.16 \times 10^{-5}$).
    \item \emph{Family-equalized aggregation:} Averaging the eight Indo-European languages into a single family mean so all nine families receive equal weight yields $9/9$ families with higher $\mathbb{E}[d]$ ($7.43\text{--}8.76$ vs.\ $6.79\text{--}7.42$), lower $\sigma(d)$ ($2.36\text{--}3.34$ vs.\ $2.85\text{--}3.63$), and lower $H(d_1, d_2)$ ($5.85\text{--}6.58$ vs.\ $6.78\text{--}7.26$ bits) relative to unrestricted shuffles, and $9/9$ families with lower $H(d_1, d_2)$ relative to C/V-preserving permutations ($5.85\text{--}6.58$ vs.\ $6.03\text{--}6.77$ bits).
\end{enumerate}

\subsection{Cross-Linguistic Differences in Peak Sharpness}
Visual inspection of Figure~\ref{fig:heatmaps} suggests that some languages (such as Yoruba, Portuguese, Turkish, Indonesian, and Spanish) exhibit a noticeably sharper, denser modal peak than others (such as Vietnamese, German, Hindi, and Finnish). To quantify this difference beyond visual impression, we computed the probability mass in the $3 \times 3$ neighborhood centered on the modal peak $(d_1^*, d_2^*)$ alongside two candidate structural predictors: the effective segment inventory size $|P|$ and the proportion of strictly alternating CVC/VCV triphones (based on PanPhon's syllabic feature \texttt{syl}).

Across the 17 languages, peak sharpness correlates significantly and positively with CVC/VCV alternation ($r = +0.616$, $p = 0.008$, Spearman $\rho = +0.498$, $p = 0.042$ for modal cell probability; $r = +0.582$, $p = 0.014$, $\rho = +0.537$, $p = 0.026$ for $3 \times 3$ peak mass), while showing a non-significant negative trend with effective inventory size ($r = -0.322$, $p = 0.208$). Languages combining moderate inventories ($|P| = 32\text{--}44$) with high CVC/VCV alternation ($61\text{--}86\%$, e.g., Yoruba 85.7\%, Indonesian 73.0\%, Turkish 70.3\%, Portuguese 63.5\%, Spanish 61.2\%) channel most transitions into C--V and V--C steps of similar feature distance, whereas heavy consonant-cluster or vowel-sequence languages (Maltese 51.0\%, Vietnamese 51.6\%, German 52.9\%, English 59.2\% CVC/VCV) and large-inventory languages with secondary contrasts such as aspiration or nasal/front-rounded vowels (Hindi $|P|=64$, Hungarian $|P|=63$, German $|P|=59$, French $|P|=51$) spread probability mass across a broader range of $(d_1, d_2)$ cells.

\section{Discussion and Limitations}
These cross-linguistic results clarify how sequential feature-space organization relates to the competing functional pressures of articulatory economy and perceptual distinctiveness introduced at the outset \citep{zipf1949,martinet1955,lindblom1990,flemming2004,kawasakifukumori1992}. Relative to unrestricted within-word permutations, natural sequences across all 17 corpora (spanning nine language families) exhibit higher expected adjacent feature distance, lower standard deviation, and lower joint triphone entropy under both Hamming and $\ell_2$-normalized cosine metrics ($W=0, p=1.53\times 10^{-5}$), clustering in an intermediate-distance band that avoids both minimal-contrast transitions (which risk perceptual confusability) and maximal feature jumps (which would require abrupt articulatory reconfiguration). Crucially, the strict within-word C/V-template-preserving control and mutual-information decomposition show how this organization is layered: coarse C/V alternation accounts for the upward shift in expected contrast ($\mathbb{E}[d]$, $p=0.051$) and overall dispersion ($\sigma(d)$, $p=0.517$), whereas natural sequences still exhibit lower joint entropy $H(d_1,d_2)$ ($16/17$ languages, $p=1.07\times 10^{-4}$) and higher mutual information $I(d_1; d_2)$ in $17/17$ languages ($W=0, p=1.53\times 10^{-5}$; $9/9$ language families). In other words, the feature-space properties of an incoming transition $d_1$ systematically constrain the outgoing transition $d_2$ beyond what C/V alternation and word-level segment inventories alone predict, a conclusion reinforced by the 16-feature non-major-class ablation ($14/17$ languages, $p=3.85\times 10^{-3}$). At the same time, because discrete feature distances do not directly measure physical articulatory displacement or acoustic discriminability, these patterns should be interpreted as distinctive-feature-space constraints consistent with economy--distinctiveness accounts, rather than direct proof of physical optimization.

Several limitations qualify these findings. First, although our expanded 17-language sample includes nine non-Indo-European languages across eight non-IE families (Uralic, Turkic, Austronesian, Koreanic, Dravidian, Afro-Asiatic, Austroasiatic, and Niger-Congo), corpus sizes vary widely from 5.6k (Tamil) and 7.1k (Yoruba) to 331k tokens (Spanish); while 17k-token subsampling and family-equalized aggregation confirm consistency within this sample, even broader sampling across Indigenous American, Sino-Tibetan, and Australian families is desirable. Second, all 17 languages rely on Epitran G2P (backed by Flite's \texttt{lex\_lookup} for English); although this places every language on a uniform methodological footing and achieves 99.7\%--100.0\% coverage, G2P transcriptions can miss lexicalized reductions, loanword adaptations, or dialect variation, and unvocalized abjads (such as Arabic) were excluded. Third, the analysis examines word-internal triphones in written UD corpora rather than continuous multi-word speech (which includes cross-word boundary transitions) or acoustic recordings, and discrete feature distances remain feature-space proxies rather than direct measures of articulatory displacement or auditory discriminability. Fourth, because we pool all word-internal triphones regardless of linear position, the present design does not test whether outgoing transition constraints strengthen directionally over the course of a word or morpheme (due to lexical cohort narrowing, affixal structure, or positional neutralization).

\section{Conclusion}
Across a diverse sample of languages from multiple families, this study shows that natural words do not arrange their speech sounds at random. Instead, adjacent segments consistently occupy a balanced middle ground in distinctive-feature space: they differ enough to stay perceptually distinct, yet avoid extreme articulatory jumps. While much of this overall spacing is driven by the familiar alternation of consonants and vowels, that coarse syllable-level pattern is not the whole story. Even when each word's consonant--vowel template and segment inventory are held fixed, consecutive transitions remain linked: the size of the feature shift entering a sound systematically constrains the shift leaving it. Together, these findings show that the classic trade-off between articulatory economy and perceptual distinctiveness shapes not only which sounds a language selects, or how it alternates consonants and vowels, but also the fine-grained sequential coordination between neighboring sounds.

Quantitatively, across our 17-language, nine-family sample, this study rejects the inventory-only null hypothesis ($H_{0,\text{unrest}}$) in all 17 corpora ($W=0, p=1.53 \times 10^{-5}$): natural segmental sequences consistently show higher expected adjacent distances ($+0.71$ to $+1.66$ features under Hamming; $+0.019$ to $+0.055$ under cosine), lower distance variability ($-0.23$ to $-1.01$ features under Hamming; $-0.002$ to $-0.035$ under cosine), and reduced joint triphone entropy ($-0.50$ to $-1.01$ bits under Hamming; $-0.29$ to $-0.74$ bits under cosine). Testing against the strict within-word C/V-template-preserving null hypothesis ($H_{0,\text{CV}}$) further shows that while $\mathbb{E}[d]$ ($p=0.051$) and $\sigma(d)$ ($p=0.517$) are not significantly separated once the word's C/V template and segment multiset are fixed, $H_{0,\text{CV}}$ is consistently rejected for joint triphone entropy ($16/17$ languages, $p=1.07 \times 10^{-4}$) and mutual information $I(d_1; d_2)$ ($17/17$ languages, $+0.007$ to $+0.101$ bits, $W=0, p=1.53 \times 10^{-5}$; $9/9$ families), demonstrating that consecutive feature transitions around a segment are sequentially coupled beyond C/V alternation. Future work employing cross-word utterance boundaries, direct acoustic measurements, position-resolved and morpheme-aware analyses (to test whether transition constraints intensify across word and morpheme unfolding), and multidimensional scaling (MDS) to embed cross-linguistic feature-transition geometries will help determine how these sub-syllabic feature-space constraints map onto physical speech production, perception, and lexical structure.

\section*{Data and Code Availability}
All Universal Dependencies v2 treebanks used in this study are publicly available from the Universal Dependencies repository (\url{https://universaldependencies.org/}). All Python analysis scripts (\texttt{analyze\_crosslinguistic.py}, \texttt{analyze\_cosine.py}, \texttt{analyze\_peak\_sharpness.py}, \texttt{analyze\_cv\_baseline.py}, \texttt{analyze\_major\_vs\_finer.py}, \texttt{run\_within\_word\_cv\_mi.py}, \texttt{run\_robustness\_checks.py}, \texttt{run\_feature\_sensitivity.py}), generated summary tables, and figures are publicly available at \url{https://github.com/leolca/triphone_feature_distance}.

\section*{Disclosure Statement}
No potential conflict of interest was reported by the author.

\section*{Declaration of Generative AI and AI-Assisted Technologies}
During the preparation of this work, the author used an AI assistant (Gemini) to assist with Python scripting for corpus processing and initial LaTeX drafting. After using this tool, the author reviewed, verified, and edited all code, statistical results, citations, and manuscript text, and takes full responsibility for the content of the published article.

\bibliographystyle{apacite}
\bibliography{paper}

\end{document}